%% file: main.tex
\documentclass[10pt, letter, onecolumn]{arxiv}

\usepackage[bibstyle=nature,citestyle=numeric-comp,natbib=true,backend=biber,maxbibnames=99,giveninits=false,sorting=none]{biblatex}
\usepackage{nameref}
\usepackage{varioref}
\usepackage{hyperref}
\usepackage[noabbrev,capitalize]{cleveref}
\usepackage{etoc}

\usepackage{graphicx}
\usepackage{pifont}
\usepackage{amssymb}
\usepackage{multirow}
\usepackage{array}
\usepackage{dblfloatfix}
\usepackage{titlesec}

\newlength{\cboxlength}
\usepackage{amsmath,xparse,mleftright}
\NewDocumentCommand{\up}{som}{%
  \IfBooleanTF{#1}
    {\upext{#3}}
    {#3\IfNoValueTF{#2}{\mathord}{#2}\uparrow}%
}
\NewDocumentCommand{\upext}{m}{%
  \mleft.\kern-\nulldelimiterspace#1\mright\uparrow
}

\usepackage{xcolor} 
\usepackage{soul} 

\usepackage{adjustbox}
\usepackage[most]{tcolorbox}
\usepackage{float}
\usepackage{xspace}
\usepackage[symbol]{footmisc}
\usepackage{lineno}

\tcbset{
  aibox/.style={
    width=394.18663pt,
    top=10pt,
    colback=white,
    colframe=black,
    colbacktitle=black,
    enhanced,
    center,
    attach boxed title to top left={yshift=-0.1in,xshift=0.15in},
    boxed title style={boxrule=0pt,colframe=white,},
  }
}
\newtcolorbox{AIbox}[2][]{aibox,title=#2,#1}

\ifdefined\red
    \renewcommand{\red}[1]{\textcolor{red}{#1}}
\else
    \newcommand{\red}[1]{\textcolor{red}{#1}}
\fi
\ifdefined\blue
    \renewcommand{\blue}[1]{\textcolor{blue}{#1}}
\else
    \newcommand{\blue}[1]{\textcolor{blue}{#1}}
\fi

\makeatletter
\renewcommand\AB@authnote[1]{}
\renewcommand\AB@affilnote[1]{}
\makeatother

\usepackage{anyfontsize} 
\usepackage{titlesec}
\titleformat{\section}{\normalfont\Large\bfseries}{\thesection}{1em}{}

\title{{\fontsize{17.5pt}{16.5pt}\selectfont A Vision-Language Foundation Model to Enhance Efficiency of Chest X-ray Interpretation}}

\author[]{Zhihong Chen$^{1,2,*}$, Maya Varma$^{1,2,3,*}$, Justin Xu$^{1,2,4}$, Magdalini Paschali$^{1,2}$, Dave Van Veen$^{1,5}$,\\Andrew Johnston$^{2}$, Alaa Youssef$^{1,2}$, Louis Blankemeier$^{1,5}$, Christian Bluethgen$^{1,6}$,\\ Stephan Altmayer$^{2}$, Jeya Maria Jose Valanarasu$^{1,3}$, Mohamed Siddig Eltayeb Muneer$^{2}$,\\Eduardo Pontes Reis$^{1,2}$, Joseph Paul Cohen$^{1}$, Cameron Olsen$^{2}$, Tanishq Mathew Abraham$^7$,\\Emily B. Tsai$^{2}$, Christopher F. Beaulieu$^{2}$, Jenia Jitsev$^{8,9}$, Sergios Gatidis$^{1,2}$,\\Jean-Benoit Delbrouck$^{1,2}$, Akshay S. Chaudhari$^{1,2,10}$, Curtis P. Langlotz$^{1,2,10,11}$}

\affil{\footnotesize{$^1$Stanford Center for Artificial Intelligence in Medicine and Imaging, Stanford University, Palo Alto, CA, USA. $^2$Department of Radiology, Stanford University, Stanford, CA, USA. $^3$Department of Computer Science, Stanford University, Stanford, CA, USA. $^4$Big Data Institute, University of Oxford, Oxford, UK. $^5$Department of Electrical Engineering, Stanford University, Stanford, CA, USA. $^6$Department of Radiology, University Hospital Zurich, Zürich, Switzerland. $^7$Stability AI, London, UK. $^8$Jülich Supercomputing Centre, Jülich, Germany.
$^9$LAION, Germany. $^{10}$Department of Biomedical Data Science, Stanford University, Stanford, CA, USA. $^{11}$Department of Medicine, Stanford University, Stanford, CA, USA. Corresponding to: \{zhihongc,mvarma2,jbdel,akshaysc,langlotz\}@stanford.edu}}

\renewcommand{\correspondingauthor}[1]{$\ast$~Equal contributions.}

\begin{document}
\begin{abstract}
Over 1.4 billion chest X-rays (CXRs) are performed annually due to their cost-effectiveness as an initial diagnostic test. This scale of radiological studies provides a significant opportunity to streamline CXR interpretation and documentation. While foundation models are a promising solution, the lack of publicly available large-scale datasets and benchmarks inhibits their iterative development and real-world evaluation. To overcome these challenges, we constructed a large-scale dataset (CheXinstruct), which we utilized to train a vision-language foundation model (CheXagent). We systematically demonstrated competitive performance across eight distinct task types on our novel evaluation benchmark (CheXbench). Beyond technical validation, we assessed the real-world utility of CheXagent in directly drafting radiology reports. Our clinical assessment with eight radiologists revealed a 36\% time saving for residents using CheXagent-drafted reports, while attending radiologists showed no significant time difference editing resident-drafted or CheXagent-drafted reports. The CheXagent-drafted reports improved the writing efficiency of both radiology residents and attending radiologists in 81\% and 61\% of cases, respectively, without loss of quality. Overall, we demonstrate that CheXagent can effectively perform a variety of CXR interpretation tasks and holds potential to assist radiologists in routine clinical workflows.
\end{abstract}

\maketitle

\begin{refsection}
\input{1_intro}
\input{2_results}
\input{3_discussion}
\input{4_figure_legends}
\input{5_method}
\clearpage
\setlength\bibitemsep{3pt}
\printbibliography
\clearpage
\input{6_appendix}
\end{refsection}
\end{document}

%% file: 1_intro.tex
\clearpage
\section*{\hspace{-0.5em}Introduction}
Chest X-rays (CXRs) are the most frequently performed imaging tests in clinical practice due to their wide availability, cost-effectiveness, and low radiation doses. CXRs comprise approximately 40\% of the 3.6 billion diagnostic X-ray examinations performed worldwide each year \supercite{paho2012world,world2016communicating,cid2024development}. Physicians obtain CXRs for diverse purposes, including diagnosing disease, monitoring longitudinal disease progression, and verifying the placement of medical devices, among others. An increasing demand for imaging studies and the subsequent interpretation and documentation of a high volume of CXRs places a significant burden on radiologists \supercite{ruutiainen2013increased,hanna2017emergency,bruls2020workload}. This can lead to burnout and may compromise diagnostic accuracy, with an increased risk of misidentification or delayed reporting of relevant findings\supercite{bhargavan2002too,lyon2015rural,rimmer2017radiologist}.

Machine learning (ML) methods have been proposed to automate the interpretation of CXRs\supercite{erickson2017machine,mcbee2018deep,rajpurkar2017chexnet,isensee2019nnu}. Traditionally, ML models have been designed with the goal of addressing a single pre-defined task, such as disease classification\supercite{rajpurkar2017chexnet,li2019deep,tiu2022chexzero}, abnormality detection\supercite{yan2018deeplesion,liu2023clip}, visual grounding\supercite{chen2023medrpg,muller2024chex}, and radiology report generation\supercite{shin2016learning,zhang2017mdnet,jing2018automatic}. Despite promising results, the capabilities of such task-specific models are restricted to a narrow scope by design. Additionally, task-specific models miss a key opportunity to leverage complementary knowledge from diverse tasks. For instance, consider the tasks of (1) radiology report generation, which involves generating a text-based radiology report given input CXR images, and (2) disease localization, which involves identifying a fine-grained region of interest (ROI) in a CXR for a specified disease. Although these tasks are noticeably distinct, training jointly on both tasks can enable a model to acquire superior capabilities, such as the ability to generate high-quality reports sensitive to fine-grained disease information.

Foundation models (FMs), a powerful class of models that can be adapted for diverse tasks, have recently emerged as a promising solution to the aforementioned challenges\supercite{bommasani2021opportunities,moor2023foundation,li2024llavamed}. In non-medical domains, FMs have demonstrated the ability to perform a range of complex reasoning and comprehension tasks\supercite{touvron2023llama,achiam2023gpt4,liu2024llava}. However, two major barriers hinder the development of FMs for CXR interpretation: (1) a lack of curated large-scale training datasets that comprise diverse tasks, and (2) the limited availability of holistic evaluation benchmarks for assessing true performance across a broad range of capabilities. Moreover, the nascent field of CXR FMs\supercite{thawkar2023xraygpt,hyland2023maira,chaves2024llavarad} has primarily focused on radiology report generation, without robust evaluation of other capabilities critical for effective CXR interpretation.

Our aim in this study was to build an FM capable of performing diverse CXR interpretation and reasoning tasks. We first collected 32 publicly available datasets and performed extensive data engineering to curate CheXinstruct, a large-scale dataset for CXR interpretation. To the best of our knowledge, CheXinstruct is the largest publicly available collection for training CXR FMs, encompassing 8.5 million training samples across 35 tasks. Next, we leveraged CheXinstruct to train CheXagent, a vision-language FM for CXR interpretation. 
We then introduced a comprehensive benchmark, CheXbench, for evaluating FMs on three image perception tasks, three image-text reasoning tasks, and two text generation tasks. CheXagent outperformed prior medical FMs, general domain FMs, and task-specific models across the evaluated tasks.

To bring CXR FMs closer to clinical readiness, we conducted a reader study with eight radiologists. 
We simulated a real-world CXR interpretation workflow, in which a radiology resident first drafts an initial radiology report; then, an attending radiologist reviews the report for accuracy and makes necessary edits. Our goal is to evaluate whether using CheXagent to draft initial radiology reports can contribute to improved CXR interpretation efficiency. Our results showed that, in comparison to residents who wrote reports from scratch, residents assisted by CheXagent-drafted reports were able to achieve an average time saving of 36\%. Additionally, we found that attending radiologists exhibited no significant time differences between editing CheXagent-drafted reports and editing resident-drafted reports, demonstrating the high quality nature of CheXagent-drafted reports. Thus, we showed that CheXagent holds potential in aiding radiologists with interpretation and documentation tasks in real-world clinical workflows. 

%% file: 2_results.tex
\section*{\hspace{-0.5em}Results}
\subsection*{Creating CheXinstruct, CheXagent, and CheXbench}
\begin{figure}[!t]
\centering
\includegraphics[width=0.95\textwidth, trim=0 0 0 0]{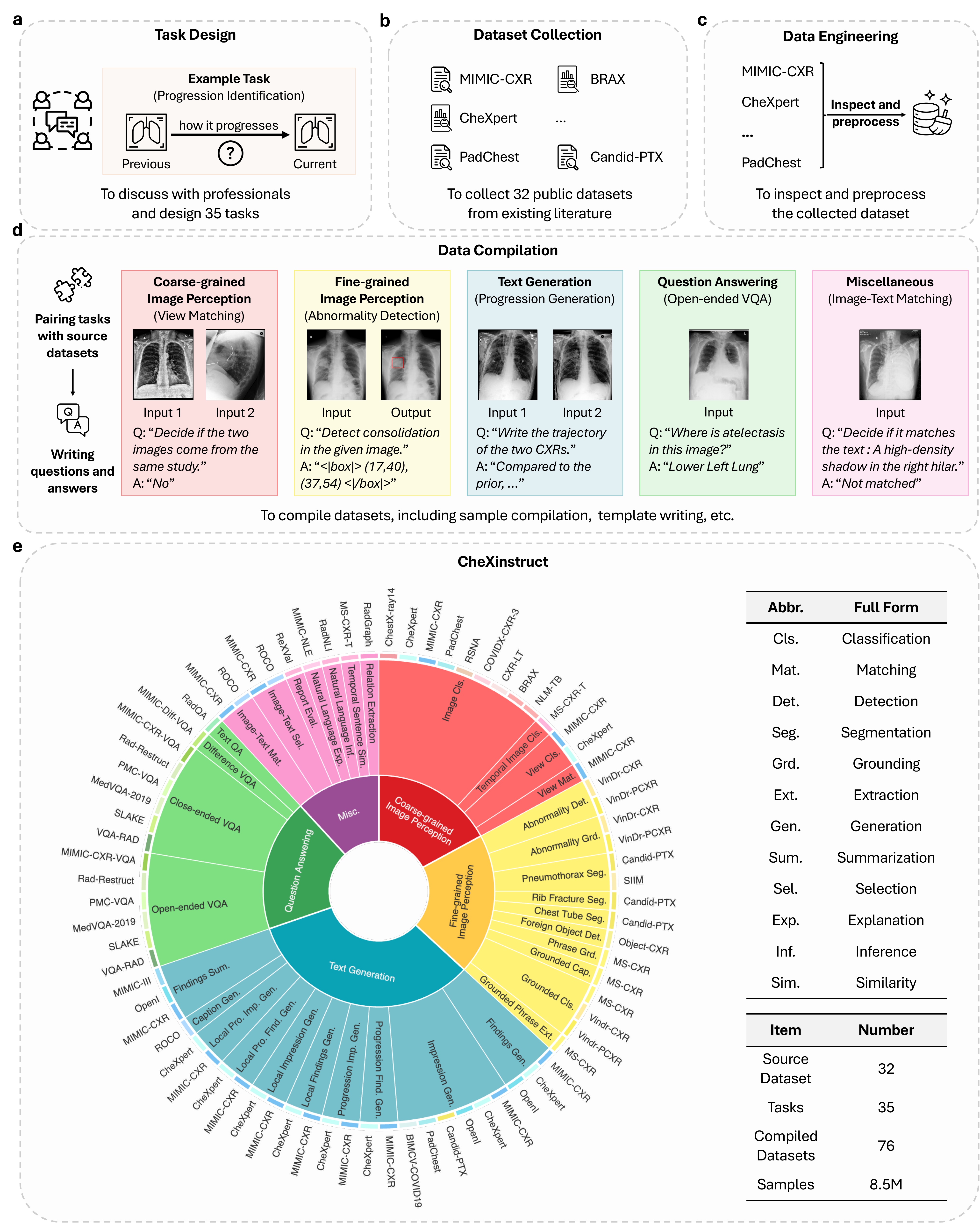}
\caption{\textbf{Curation of CheXinstruct.} a, Identification of CXR interpretation tasks. We defined 35 tasks that users are likely to perform with CXR FMs. b, Source dataset collection. To create training data samples for each of our defined tasks, we collected 32 public datasets. c, Data engineering. We performed both manual quality control and automated data engineering to preprocess collected source data. d, CheXinstruct compilation. We used the preprocessed datasets to generate training samples for each of our 35 defined tasks. e, Overview of CheXinstruct with data statistics.}
\label{fig:chexinstruct}
\end{figure}
For an FM to perform diverse CXR interpretation and reasoning tasks, it must effectively interact with diverse input queries. This necessitates a large and diverse training dataset with data triplets consisting of plausible queries (referred to as \textit{instructions}), images, and desired model responses. To build such a dataset, we first defined a series of 35 tasks (Fig.~\ref{fig:chexinstruct}a). Each task requires either (i) the ability to perceive and understand visual characteristics of a CXR (\textit{e.g.,} the view matching task, where the goal is to determine whether two CXR views are from the same imaging study) or (ii) the ability to make reasonable inferences and clinical decisions from a given CXR (\textit{e.g.,} open-ended visual question-answering (VQA), where the goal is to answer a free-form question about a provided CXR).
To generate training data associated with each task, we collected 32 publicly available source datasets (Fig.~\ref{fig:chexinstruct}b). 
We performed extensive manual and automated data engineering on the source datasets to verify quality and unify their diverse structures (Fig.~\ref{fig:chexinstruct}c).
We then paired each task with various source datasets, using the engineered annotations to construct instruction-response pairs (Fig.~\ref{fig:chexinstruct}d). The final training dataset, referred to as \textit{CheXinstruct}, consists of 8.5 million data triplets, each with an instruction, a response, and at least one image (Fig.~\ref{fig:chexinstruct}e). We note that some tasks (\textit{e.g.}, the findings summarization task) included in CheXinstruct are text-only, in which case no images are included in the triplet.

\begin{figure}[!t]
\centering
\includegraphics[width=0.95\textwidth, trim=0 0 0 0]{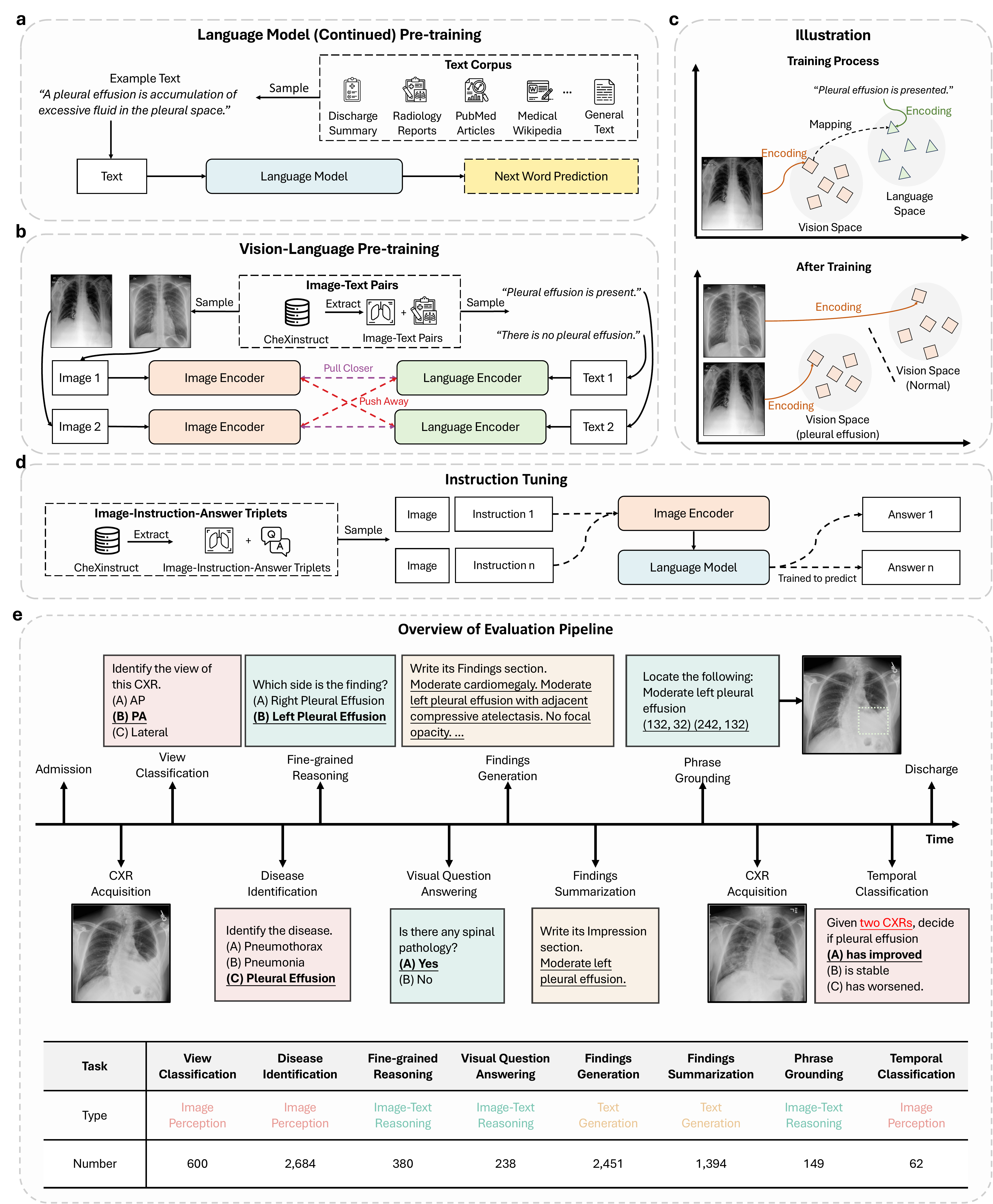}
\caption{\textbf{Training and evaluating CheXagent}. a, To develop CheXagent, we first trained a language model on clinical text. b, We then trained an image encoder to learn useful visual representations of imaging findings by leveraging paired text. c, This procedure enabled the visual encoder to capture semantic meaning with respect to key findings within its latent representation space. d, Finally, we jointly trained the image encoder and language model on data triplets from CheXinstruct, providing CheXagent with the capability to respond to user instructions. e, We constructed eight evaluation tasks to assess image perception, reasoning, and text generation capabilities.}
\label{fig:chexagent}
\end{figure}
We utilized CheXinstruct to train \textit{CheXagent} (Fig~\ref{fig:chexagent}), an FM that takes images and an instruction as input and generates a response to complete the instruction. 
CheXagent is composed of an image encoder for interpreting CXRs and a large language model for understanding and generating text. The image encoder divides each image into patches and computes a representation for each patch; then, the language model processes these patch representations alongside the instruction and generates a response.
We trained CheXagent using a three-stage process. First, the language model was trained on clinical text (discharge summaries, radiology reports, clinical guidelines, and medical articles) with the goal of acquiring broad medical knowledge (Fig.~\ref{fig:chexagent}a).
Then, the image encoder was trained using SigLIP\supercite{zhai2023siglip}, an approach that aims to learn useful representations of imaging findings guided by their textual descriptions. This was achieved by teaching the model to match correct image-text pairs while simultaneously distinguishing them from incorrect pairings. This training stage utilizes CXRs and their paired radiology reports (Fig.~\ref{fig:chexagent}b) and enables the image encoder to capture semantic meaning within its representation space (illustrated in Fig.~\ref{fig:chexagent}c).
Finally, the image encoder and language model were trained jointly using the data triplets in the CheXinstruct dataset; this stage enables the model to learn how to respond to instructions across a variety of diverse CXR interpretation tasks (Fig.~\ref{fig:chexagent}d).

We developed an evaluation benchmark, \textit{CheXbench}, to assess the capabilities of FMs in interpreting CXRs (Fig.~\ref{fig:chexagent}e). Specifically, we evaluated the ability of FMs to understand the visual content of CXRs (\textit{image perception}), perform complex reasoning tasks on CXRs (\textit{image-text reasoning}), and generate and understand clinical text. For each evaluation task, we formatted the task as an instruction; then, we provided the instruction and corresponding image(s) as input to CheXagent and evaluated the quality of the generated response.

\subsection*{Performance on Image Perception}
\begin{figure}[!t]
\centering
\includegraphics[width=0.99\textwidth, trim=0 0 0 0]{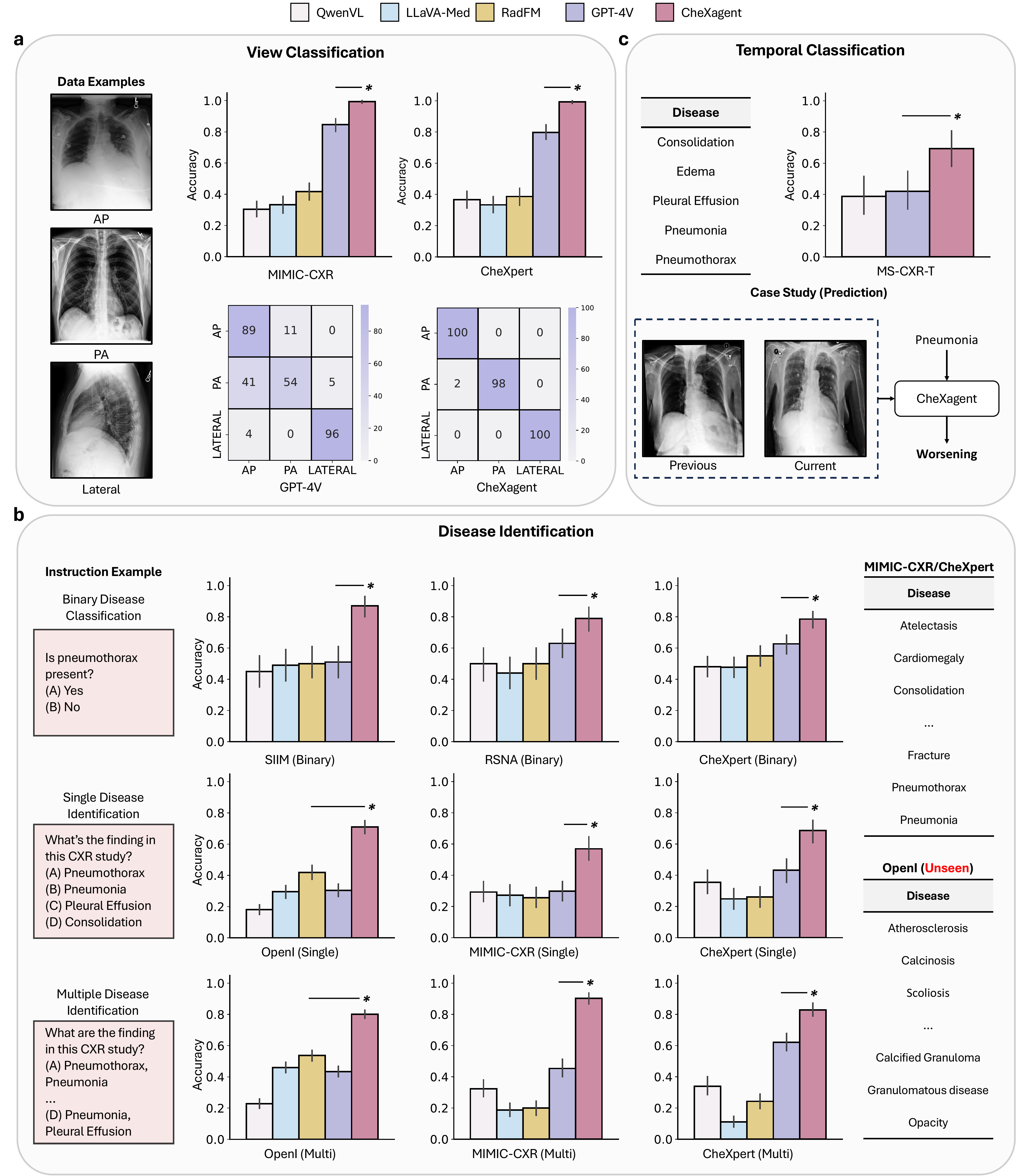}
\caption{\textbf{Technical evaluation on image perception tasks.} a, Performance of FMs on view classification. Bar graphs show mean accuracy with 95\% confidence intervals.  Confusion matrices compare predictions of CheXagent and GPT-4V. b, Performance of FMs on disease identification with three subtasks.  Bar graphs show mean accuracy with 95\% confidence intervals. Evaluations on OpenI, which was unseen during CheXagent training, assess generalization capabilities. c, Performance of FMs on temporal classification. The bar graph shows mean accuracy with 95\% confidence intervals. We provide one example of a prediction generated by CheXagent on the temporal classification task.}
\label{fig:evaluation-image-perception}
\end{figure}
We first evaluated the ability of CheXagent to understand the visual content of CXRs (Fig.~\ref{fig:evaluation-image-perception}). We refer to this evaluation axis as \textit{image perception}. We assessed image perception capabilities with three tasks: (1) View Classification, which involves classifying the imaging view of a CXR; (2) Disease Identification, which involves identifying key findings in a CXR; and (3) Temporal Classification, which involves classifying the progression of a disease between two CXR studies obtained at different times. We compared CheXagent with an open general-domain vision-language model (QwenVL\supercite{bai2023qwen}), two medical-domain vision-language models (LLaVA-Med\supercite{li2024llavamed} and RadFM\supercite{wu2023radfm}), and a proprietary model (GPT-4V\supercite{achiam2023gpt4}). We formatted each task as an instruction with multiple choices; then, we evaluated the accuracy of each FM in generating the correct response. Our results demonstrated that CheXagent consistently outperformed other FMs across all three tasks.

On the task of View Classification (Fig.~\ref{fig:evaluation-image-perception}a), CheXagent achieved an accuracy of 0.993 (95\%CI=0.983-1.000) on MIMIC-CXR\supercite{johnson2019mimiccxr} and 0.993 (95\%CI=0.983-1.000) on CheXpert\supercite{irvin2019chexpert}. Among the baseline models, GPT-4V performed the best with accuracies of 0.847 (95\%CI=0.807-0.887) on MIMIC-CXR and 0.797 (95\%CI=0.750-0.840) on CheXpert. The confusion matrix indicated that while GPT-4V can distinguish front and lateral views well, it struggles to differentiate between AP and PA frontal views.

On the task of Disease Identification (Fig.~\ref{fig:evaluation-image-perception}b), we evaluated models using three subtasks: (1) Binary Disease Classification, which involves identifying the presence or absence of a finding; (2) Single Disease Identification, which involves identifying a single finding present in a CXR given four options; and (3) Multiple Disease Identification, which involves identifying a set of multiple findings present in a CXR given four options. On the subtask of Binary Disease Classification, CheXagent achieved an accuracy of 0.870 (95\%CI=0.800-0.930) for pneumothorax recognition on the SIIM\supercite{siim} dataset, 0.790 (95\%CI=0.710-0.860) for pneumonia recognition on the RSNA\supercite{shih2019rsna} dataset, and 0.785 (95\%CI=0.734-0.841) for various diseases on the CheXpert dataset\supercite{irvin2019chexpert}. On the subtask of Single Disease Identification, CheXagent achieved an accuracy of 0.710 (95\%CI=0.672-0.750) on the OpenI\supercite{demner2016openi} dataset, 0.569 (95\%CI=0.497-0.636) on the MIMIC-CXR\supercite{johnson2019mimiccxr} dataset, and 0.686 (95\%CI=0.621-0.751) on the CheXpert dataset. On the subtask of Multiple Disease Identification, CheXagent achieved promising performance with accuracies of 0.800 (95\%CI=0.773-0.827), 0.903 (95\%CI=0.870-0.933), and 0.829 (95\%CI=0.782-0.868) on OpenI, MIMIC-CXR, and CheXpert, respectively. Notably, the OpenI dataset was entirely held out during the training of CheXagent. As demonstrated in Fig.~\ref{fig:evaluation-image-perception}b, OpenI has a labeling scheme that differs from MIMIC-CXR and CheXpert, providing evidence that CheXagent effectively generalizes to out-of-distribution images and disease labels.

On the task of Temporal Classification (Fig.~\ref{fig:evaluation-image-perception}c), CheXagent achieved an accuracy of 0.694 (95\%CI=0.565-0.790) on MS-CXR-T \supercite{boecking2022mscxr}, outperforming the other evaluated FMs (0.387 (95\%CI=0.258-0.500) for QwenVL and 0.419 (95\%CI=0.306-0.548) for GPT-4V). In Fig.~\ref{fig:evaluation-image-perception}c, we provided an example demonstrating CheXagent's assessment of pneumonia progression. Our results demonstrate the ability of CheXagent to process multiple CXR studies and understand temporal patterns.

\subsection*{Performance on Image-Text Reasoning}
\begin{figure}[!t]
\centering
\includegraphics[width=0.99\textwidth, trim=0 0 0 0]{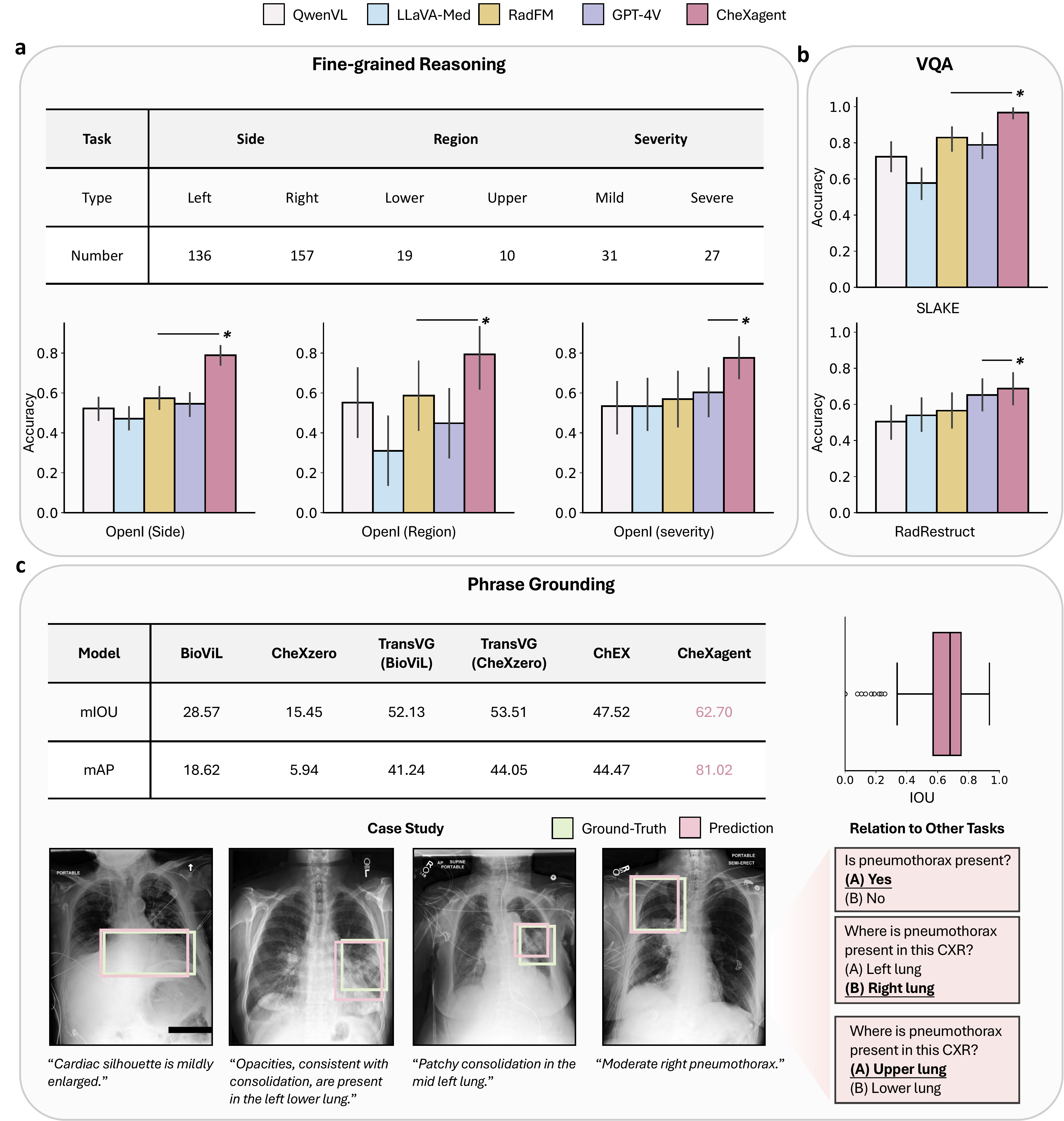}
\caption{\textbf{Technical evaluation on image-text reasoning tasks}. a, Performance of FMs on fine-grained reasoning. We provide the number of samples included in each subtask. Bar graphs show mean accuracy with 95\% confidence intervals. b, Performance of FMs on visual question-answering (VQA). Bar graphs show mean accuracy with 95\% confidence intervals. c, Performance on phrase grounding. We report mean intersection over union (mIOU) and mean average precision (mAP) scores. The box plot shows the distribution of IOU scores for CheXgaent. We also provide several examples comparing bounding boxes predicted by CheXagent to ground truth localizations. Lastly, we provide an example that relates the task of phrase grounding to VQA; users can iteratively ask questions to CheXagent in order to roughly ground findings in an image.}
\label{fig:evaluation-image-text-reasoning}
\end{figure}
Next, we evaluated the ability of CheXagent to perform joint reasoning over images and text. We refer to this evaluation axis as \textit{image-text reasoning}. 
We assessed image-text reasoning capabilities with three tasks: (1) Fine-Grained Reasoning, which evaluates the ability of a model to differentiate between two subtly different findings; (2) Visual Question Answering, which involves answering open-ended free-form questions about the content of a CXR; and (3) Phrase Grounding, which involves localizing the region in a CXR corresponding to a specific sentence from a radiology report. For Fine-Grained Reasoning and Visual Question Answering, we formatted the task as an instruction with multiple choices and evaluated the accuracy of FMs in generating the correct response. For Phrase Grounding, we evaluated the accuracy of bounding box coordinates generated by the FM in its response. Our results demonstrated that CheXagent consistently outperforms other FMs and task-specific models.

On the task of Fine-Grained Reasoning (Fig.~\ref{fig:evaluation-image-text-reasoning}a), we evaluated the ability of models to differentiate whether a finding is (1) located on the left or right side of the body (\textit{side}), (2) located on the lower or upper region of the lung (\textit{region}), and (3) mild or severe in presentation (\textit{severity}). CheXagent achieved an accuracy of 0.788 (95\%CI=0.737-0.836) on the side subtask, 0.793 (95\%CI=0.655-0.931) on the region subtask, and 0.776 (95\%CI=0.672-0.879) on the severity subtask. We note that the OpenI dataset was held out from training, and CheXagent was not specifically optimized for this task; this demonstrates the generalization capabilities of CheXagent.

On the task of Visual Question Answering (VQA) (Fig.~\ref{fig:evaluation-image-text-reasoning}b), we evaluated models on VQA samples derived from the SLAKE\supercite{liu2021slake} and RadRestruct\supercite{pellegrini2023radrestruct} datasets, with the latter held out during training. CheXagent outperformed the baseline models, achieving an accuracy of 0.967 (95\%CI=0.935-0.992) on SLAKE and 0.687 (95\%CI=0.600-0.774) on RadRestruct.

On the task of Phrase Grounding (Fig.~\ref{fig:evaluation-image-text-reasoning}c), CheXagent achieved a mean intersection over union (mIOU) score of 0.627 and a mean average precision (mAP) score of 0.810, outperforming four previously-developed approaches: two zero-shot contrastive models (BioViL\supercite{bannur2023biovil} and CheXzero\supercite{tiu2022chexzero}), one single-task supervised visual grounding model (TransVG\supercite{deng2021transvg}), and one multi-task supervised model (ChEX\supercite{muller2024chex}). In particular, we note that CheXagent outperformed task-specific models on Phrase Grounding, suggesting that CheXagent effectively learned complementary knowledge from the diverse CheXinstruct tasks.

\subsection*{Performance on Text Generation}
\begin{figure}[!t]
\centering
\includegraphics[width=0.99\textwidth, trim=0 0 0 0]{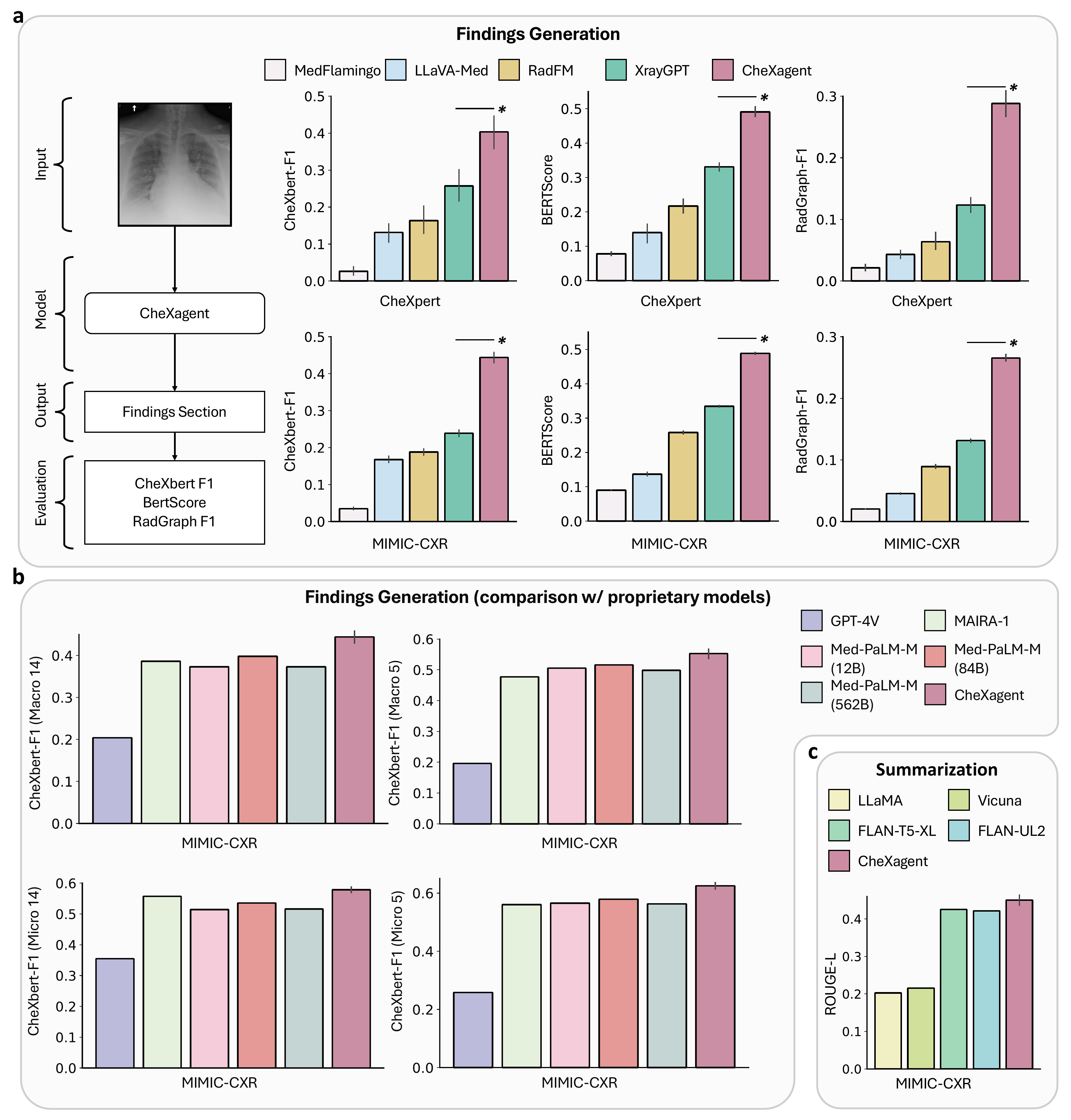}
\caption{\textbf{Technical evaluation of text generation tasks}. a, Comparisons of CheXagent with publicly-available medical FMs on findings generation.  We evaluate across two datasets (MIMIC-CXR and CheXpert). Bar graphs show mean CheXbert-F1, BERTScore, and RadGraph-F1 scores with 95\% confidence intervals. b, Comparisons of CheXagent with proprietary FMs on findings generation.  We evaluate on the MIMIC-CXR dataset. Bar graphs show mean CheXbert-F1 scores, with 95\% confidence intervals reported for CheXagent. c, Performance of large language models on findings summarization. The bar graph shows mean ROUGE-L scores, with 95\% confidence intervals reported for CheXagent.}
\label{fig:evaluation-text-generation}
\end{figure}
We evaluated the ability of CheXagent to generate and understand clinical text (Fig.~\ref{fig:evaluation-text-generation}) with two tasks: (1) Findings Generation, which involves generating the Findings section of a radiology report given at least one CXR, and (2) Findings Summarization, which involves generating the Impressions section of a radiology report given the Findings section. We compared CheXagent with a variety of FMs, including publicly available and proprietary models. Our results demonstrate that CheXagent achieves competitive performance.

On the task of Findings Generation, we evaluated models on two datasets (MIMIC-CXR\supercite{johnson2019mimiccxr} and CheXpert\supercite{irvin2019chexpert}) using three evaluation metrics (CheXbert-F1\supercite{miura2021ifcc}, BERTScore\supercite{zhangbertscore}, and RadGraph-F1\supercite{yu2023radgraphf1,delbrouck2022radgraphf1}). CheXagent achieved superior performance compared to publicly available baselines, attaining a CheXbert-F1\supercite{miura2021ifcc} score of 0.403 (95\%CI=0.356-0.448), a BERTScore\supercite{zhangbertscore} score of 0.491 (95\%CI=0.475-0.507), and a RadGraph-F1\supercite{yu2023radgraphf1,delbrouck2022radgraphf1} score of 0.288 (95\%CI=0.266-0.310) on the CheXpert dataset, and a CheXbert-F1 score of 0.444 (95\%CI=0.428-0.460), a BERTScore of 0.488 (95\%CI=0.484-0.493), and a RadGraph-F1 score of 0.266 (95\%CI=0.260-0.272) on the MIMIC-CXR dataset(Fig.~\ref{fig:evaluation-text-generation}a). Additionally, we compared CheXagent with proprietary models, including MAIRA-1\supercite{hyland2023maira}, Med-PaLM-M\supercite{tu2024medpalmm}, and GPT-4V\supercite{achiam2023gpt4}, using the CheXbert-F1 metric. We observed that CheXagent outperformed proprietary models in all four variants of the CheXbert-F1 score (Fig.~\ref{fig:evaluation-text-generation}b).

On the task of Findings Summarization, CheXagent achieved performance competitive with baseline models (LLaMA\supercite{touvron2023llama}, Vicuna\supercite{vicuna2023}, FLAN-T5-XL\supercite{chung2024flan}, and FLAN-UL2\supercite{chung2024flan}), achieving a ROUGE-L score (a classic text summarization metric) of 0.450 (95\%CI=0.435-0.465). This demonstrated the ability of CheXagent to effectively perform text-only tasks (Fig.~\ref{fig:evaluation-text-generation}c).

\subsection*{Clinical Evaluation: Reader Study}
\begin{figure}[!t]
\centering
\includegraphics[width=0.96\textwidth, trim=0 0 0 0]{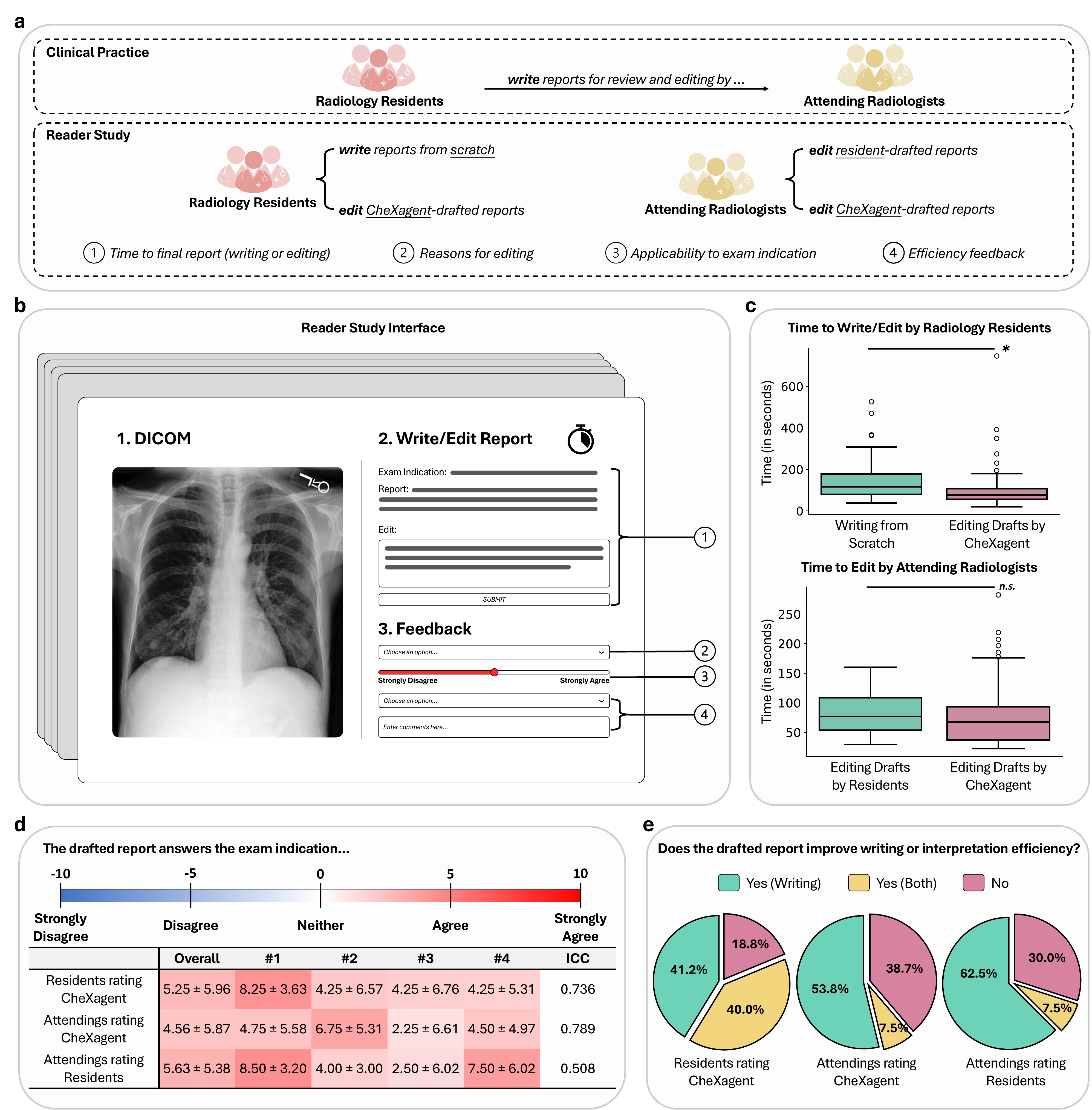}
\caption{\textbf{Clinical reader study.} a, Overview of study design. Our reader study was designed to parallel real-world academic clinical workflows, where radiology residents draft initial radiology reports and attending radiologists make necessary edits. In our study, we compared settings where radiology residents wrote reports from scratch with settings where radiology residents edited reports drafted by CheXagent. We also compared settings where attending radiologists edited reports written by residents with settings where attending radiologists edited reports drafted by CheXagent. We collected data on time required to produce a final report, applicability of the report to the exam indication, radiologists' reasons for editing reports, and radiologists' opinions on whether CheXagent-drafted reports helped with improving interpretation or writing efficiency. b, Reader study interface. For each study, readers were presented with the CXR(s) in DICOM format, exam indication, and a drafted report if applicable \raisebox{.5pt}{\textcircled{\raisebox{-.9pt} {1}}}. Fields were provided to collect feedback on reasons for editing drafted reports \raisebox{.5pt}{\textcircled{\raisebox{-.9pt} {2}}}, applicability to exam indication \raisebox{.5pt}{\textcircled{\raisebox{-.9pt} {3}}}, and efficiency \raisebox{.5pt}{\textcircled{\raisebox{-.9pt} {4}}}. c, Distributions of the time (in seconds) to required to produce a final report for residents (top) and attendings (bottom). Asterisk (*) denotes statistical significance with a two-sided Mann-Whitney U test, $P < 0.0001$; \textit{n.s.} denotes differences that are not statistically significant. d, Evaluations on whether drafted reports answer the initial exam indication. Radiologists score reports on a five-point Likert scale ranging from -10 to 10. e, Opinions of radiologists on whether drafted reports improved their report writing and/or CXR interpretation efficiency.}
\label{fig:reader-study}
\end{figure}
We evaluated the utility of CheXagent in clinical settings by conducting a reader study. In clinical workflows in academic practice, the process of interpreting a CXR study typically involves two steps. First, a radiology resident interprets the provided CXR study and drafts an initial radiology report; then, an attending radiologist reviews the report for accuracy and make any necessary edits (Fig.~\ref{fig:reader-study}a).

Our reader study focused on the role of CheXagent in drafting initial radiology reports (Fig.~\ref{fig:reader-study}a and b). We quantitatively assessed (1) whether using CheXagent-drafted reports improves radiologist efficiency and (2) whether CheXagent-drafted reports accurately address the reason for the exam (exam indication). Additionally, we collected feedback from readers with respect to (1) the quality of the CheXagent-drafted reports and (2) the effects of CheXagent-drafted reports on radiologist efficiency. Eight radiologists, including four resident radiologists and four attending radiologists, participated in our reader study.

We first quantitatively evaluated whether CheXagent-drafted reports improve radiologist efficiency. We compared the time for radiology residents to edit a CheXagent-drafted report with the time to draft an initial radiology report from scratch (Fig.~\ref{fig:reader-study}c). Across four radiology residents, we observed significant time savings when using CheXagent-drafted reports ($99.9\pm97.3$ seconds vs. $156.4\pm115.9$ seconds; $p < 0.0001$). We then compared whether the time taken for attending radiologists to review and edit a CheXagent-drafted report\footnote[3]{Here, the report was drafted by CheXagent only and not reviewed or modified by radiology residents.} was similar to the time to review and edit a resident-drafted report. Across four attending radiologists, we observed that the elapsed times were comparable ($79.7\pm54.6$ seconds vs. $83.0\pm36.3$ seconds; $p > 0.1$). 

Next, we quantitatively evaluated the effectiveness of CheXagent-drafted reports in addressing the exam indication (Fig.~\ref{fig:reader-study}d). Radiology residents largely agreed that reports drafted by CheXagent accurately addressed the exam indication, with a rating of $5.25\pm5.96$ on a 5-point Likert scale weighted between -10 and 10. Attending radiologists found that both resident-drafted and CheXagent-drafted reports addressed the exam indication, with mean ratings of $5.63\pm5.38$ and $4.56\pm5.87$, respectively; here, \textit{no} significant difference was observed ($p > 0.1$), demonstrating the high quality of CheXagent-drafted reports.
We then computed agreement ratios, defined as the proportion of cases where the reader `agrees' or `strongly agrees' that the drafted report answers the exam indication. We observed agreement ratios of 0.788 for radiology residents and 0.738 for attending radiologists when rating CheXagent-drafted reports. We also demonstrated the reliability of scores across readers in the study, with moderate to high interrater correlation coefficients (ICC).

We collected feedback from readers with respect to the quality of the CheXagent-drafted reports; in particular, we asked readers to provide their reasons for any edits made to the CheXagent-drafted reports. We found that 52.5\% of reports were modified by residents due to the report content, such as missing or false predictions and misassessment of finding severities. 32.5\% of reports were edited due to style. The corresponding numbers for attending radiologists are 51.3\% and 27.5\% for report content and style, respectively.

We also collected qualitative feedback on how CheXagent-drafted reports affected both CXR interpretation and report writing efficiency (Fig.~\ref{fig:reader-study}e). Residents reported that using a CheXagent-drafted report improved report writing efficiency in 81.2\% of cases. Residents also found that nearly half of these cases improved both writing and interpretation efficiency. For attending radiologists, both CheXagent-drafted and resident-drafted reports contributed to improved CXR interpretation efficiency in few cases (7.5\% of cases for both). However, attending radiolgists improved report writing efficiency in over half of all cases (61.3\% of cases with a CheXagent-drafted report and 70.0\% of cases with a resident-written report). In Extended Data Fig.~\ref{fig:reader-study-case-study}, we provided examples of cases by CheXagent (reviewed by radiology residents or attending radiologists) where (1) CheXagent contributed to improved CXR interpretation and writing efficiency (23.8\% of all the cases), (2) CheXagent improved \textit{only} writing efficiency (47.5\%), and (3) CheXagent did not improve efficiency (28.8\% ).

Ultimately, the results of our reader study demonstrates that CheXagent can improve clinical workflows. In particular, CheXagent holds potential to serve as a copilot for radiologists to improve reporting efficiency. Evaluations by attending radiologists also confirmed the quality and utility of CheXagent-drafted reports.

%% file: 3_discussion.tex
\section*{\hspace{-0.5em}Discussion}
In this study, we developed and evaluated CheXagent, a vision-language FM capable of performing diverse CXR interpretation tasks. To train CheXagent, we curated CheXinstruct; to the best of our knowledge, CheXinstruct is the largest and most diverse CXR FM training dataset to date, with 8.5 million training data samples from 32 publicly available source datasets. Our evaluations on our novel benchmark CheXbench demonstrated that CheXagent is capable of (1) understanding the visual content of CXRs, (2) performing complex reasoning tasks on CXRs, (3) generating and understanding clinical text, and (4) aiding in real-world clinical settings.

Several FMs have been introduced recently to automate CXR interpretation\supercite{thawkar2023xraygpt,hyland2023maira,tu2024medpalmm}, focusing predominantly on radiology report generation. 
This work aims to build a model capable of performing perception and reasoning tasks that extend beyond radiology report generation. To this end, our evaluations demonstrated that CheXagent is capable of identifying CXR imaging views, monitoring longitudinal disease progression, classifying diseases and critical findings, reasoning through fine-grained queries, performing visual question-answering, and localizing findings to corresponding image regions. Across the evaluations, we observed that CheXagent consistently outperforms baselines, including significantly larger FMs like GPT-4V. The strong performance of CheXagent across these tasks can be attributed to the use of the CheXinstruct dataset for model training. CheXinstruct consists of data triplets with instructions, images, and desired responses across 35 distinct tasks; training with such a dataset enabled CheXagent to acquire diverse capabilities and perform any CXR interpretation task at inference time simply by framing the task as a multiple-choice or open-ended instruction. This represents a significant advancement in comparison to traditional task-specific approaches, where new models must be developed for each task of interest.

In particular, our evaluations on CheXbench highlighted the reasoning capabilities of CheXagent. We specifically introduced the fine-grained reasoning task in CheXbench in order to evaluate the extent to which FMs can distinguish between subtly different findings, such as ``left-sided pleural effusion'' and ``right-sided pleural effusion''. Performing this task requires the ability to perform spatial and compositional reasoning, a skill that is often trivial for humans but challenging for vision-language models as shown in prior works\supercite{ma2023crepe,wangdiagnosing}.
We note that examples of fine-grained reasoning tasks are \textit{not} explicitly included in the CheXinstruct dataset, making this an out-of-distribution evaluation. Regardless, CheXagent demonstrated strong performance, suggesting that leveraging complementary knowledge from diverse tasks during training can improve performance on unseen tasks. CheXagent further demonstrated spatial and compositional reasoning abilities with strong performance on the phrase grounding task, which involves localizing a phrase or sentence to a corresponding region of a CXR. Whereas many existing FMs are incapable of generating bounding box coordinates for such a task, CheXagent yielded highly accurate bounding boxes and outperformed multiple task-specific approaches and FMs.

In addition to generalizing to out-of-distribution tasks like fine-grained reasoning, CheXagent also generalized to out-of-distribution datasets. Recent works have suggested that medical AI models trained on data from a single institution often fail to generalize to data from other institutions, likely due to models overfitting to the training distribution or relying heavily on spurious features\supercite{rueckel2020impact}. In order to mitigate this issue, we designed CheXinstruct to incorporate 32 publicly available datasets collected from diverse countries and institutions. We evaluated the ability of CheXagent to generalize to out-of-distribution data by excluding samples from the OpenI dataset during training; OpenI, which includes a labeling schema that differs substantially from other datasets, was used solely for evaluation purposes in this work. Through the disease identification and fine-grained reasoning tasks, we demonstrated that CheXagent can effectively generalize to unseen data. On the task of disease identification, performance trends on OpenI closely mirrored those on MIMIC-CXR and CheXpert. This suggests that the diverse datasets included in CheXinstruct prevented CheXagent from overfitting to a single data distribution and can enable effective generalization.

Prior works\supercite{thawkar2023xraygpt,li2024llavamed} on medical FMs predominantly evaluated model quality with automated metrics. However, recent studies\supercite{van2024adapted} have suggested that automated evaluations may not be sufficient, particularly when analyzing complex clinical text generated by models. Rather, evaluations by expert physicians are critical for assessing the utility of FMs in real-world clinical environments. To this end, we conducted a reader study with eight radiologists in order to rigorously evaluate the utility of CheXagent in clinical settings. We demonstrated that using CheXagent to draft radiology reports rather than writing from scratch can contribute to significant efficiency benefits (36\% time savings for radiology residents) while maintaining quality. Our results suggest that CheXagent can assist with reducing the substantial interpretation and documentation burdens placed on radiologists. 

Our study presents several opportunities for future work. First, CheXagent is a lightweight FM with 3.1 billion parameters, which presents several advantages, such as fast inference time and a low GPU memory footprint; however, evidence in the non-medical domain has suggested that larger models tend to result in stronger performance. Future work can explore the effects of model scaling laws in the context of CXR FMs. In addition, due to the ability of CheXagent to accurately perform multiple tasks, it can serve as a foundation to develop an autonomous agent\supercite{franklin1996agent,yao2022react,yao2024tree} for robust interpretation. For instance, CheXagent could be further enhanced by executing self-improvement\supercite{huang2023selfimprove,yuanself} loops, iteratively improving upon the performance by validating its own generations or synthesizing new training data. Furthermore, there are opportunities to expand the scope of our clinical reader study in the future. In particular, the use of AI tools in clinical settings may have impacts on medical student and resident education. Future studies can evaluate how AI copilots with radiology report writing capabilities enhance or detract from medical education. Future clinical studies can also compare CheXagent-drafted reports with reports dictated by radiologists using automated speech recognition, rather than those typed by hand.

Ultimately, we presented an FM capable of improving CXR interpretation efficiency while maintaining quality, as demonstrated by comprehensive evaluations across diverse tasks and a reader study with expert radiologists. Our large-scale training dataset, CheXinstruct, can enable the research and development of future FMs, and our proposed benchmark, CheXbench, can allow for standardized evaluation of future FMs on CXR interpretation tasks. Our work provides a foundation for further research into the integration and potential impact of FMs in clinical practice.

%% file: 4_figure_legends.tex
\clearpage
\section*{\hspace{-0.5em}Figure Legends}
\underline{Figure 1}

\textbf{Curation of CheXinstruct.} a, Identification of CXR interpretation tasks. We defined 35 tasks that users are likely to perform with CXR FMs. b, Source dataset collection. To create training data samples for each of our defined tasks, we collected 32 public datasets. c, Data engineering. We performed both manual quality control and automated data engineering to preprocess collected source data. d, CheXinstruct compilation. We used the preprocessed datasets to generate training samples for each of our 35 defined tasks. e, Overview of CheXinstruct with data statistics.

\underline{Figure 2}

\textbf{Training and evaluating CheXagent}. a, To develop CheXagent, we first trained a language model on clinical text. b, We then trained an image encoder to learn useful visual representations of imaging findings by leveraging paired text. c, This procedure enabled the visual encoder to capture semantic meaning with respect to key findings within its latent representation space. d, Finally, we jointly trained the image encoder and language model on data triplets from CheXinstruct, providing CheXagent with the capability to respond to user instructions. e, We constructed eight evaluation tasks to assess image perception, reasoning, and text generation capabilities.

\underline{Figure 3}

\textbf{Technical evaluation on image perception tasks.} a, Performance of FMs on view classification. Bar graphs show mean accuracy with 95\% confidence intervals.  Confusion matrices compare predictions of CheXagent and GPT-4V. b, Performance of FMs on disease identification with three subtasks.  Bar graphs show mean accuracy with 95\% confidence intervals. Evaluations on OpenI, which was unseen during CheXagent training, evaluate generalization capabilities. c, Performance of FMs on temporal classification. The bar graph shows mean accuracy with 95\% confidence intervals. We provide one example of a prediction generated by CheXagent on the temporal classification task.

\underline{Figure 4}

\textbf{Technical evaluation on image-text reasoning tasks}. a, Performance of FMs on fine-grained reasoning. We provide the number of samples included in each subtask. Bar graphs show mean accuracy with 95\% confidence intervals. b, Performance of FMs on visual question-answering (VQA). Bar graphs show mean accuracy with 95\% confidence intervals. c, Performance on phrase grounding. We report mean intersection over union (mIOU) and mean average precision (mAP) scores. The box plot shows the distribution of IOU scores for CheXgaent. We also provide several examples comparing bounding boxes predicted by CheXagent to ground truth localizations. Lastly, we provide an example that relates the task of phrase grounding to VQA; users can iteratively ask questions to CheXagent in order to roughly ground findings in an image.

\underline{Figure 5}

\textbf{Technical evaluation of text generation tasks}. a, Comparisons of CheXagent with publicly-available medical FMs on findings generation.  We evaluate across two datasets (MIMIC-CXR and CheXpert). Bar graphs show mean CheXbert-F1, BERTScore, and RadGraph-F1 scores with 95\% confidence intervals. b, Comparisons of CheXagent with proprietary FMs on findings generation.  We evaluate on the MIMIC-CXR dataset. Bar graphs show mean CheXbert-F1 scores, with 95\% confidence intervals reported for CheXagent. c, Performance of large language models on findings summarization. The bar graph shows mean ROUGE-L scores, with 95\% confidence intervals reported for CheXagent.

\underline{Figure 6}

\textbf{Clinical reader study.} a, Overview of study design. Our reader study was designed to parallel real-world academic clinical workflows, where radiology residents draft initial radiology reports and attending radiologists make necessary edits. In our study, we compared settings where radiology residents wrote reports from scratch with settings where radiology residents edited reports drafted by CheXagent. We also compared settings where attending radiologists edited reports written by residents with settings where attending radiologists edited reports drafted by CheXagent. We collected data on time required to produce a final report, applicability of the report to the exam indication, radiologists' reasons for editing reports, and radiologists' opinions on whether CheXagent-drafted reports helped with improving interpretation or writing efficiency. b, Reader study interface. For each study, readers were presented with the CXR(s) in DICOM format, exam indication, and a drafted report if applicable \raisebox{.5pt}{\textcircled{\raisebox{-.9pt} {1}}}. Fields were provided to collect feedback on reasons for editing drafted reports \raisebox{.5pt}{\textcircled{\raisebox{-.9pt} {2}}}, applicability to exam indication \raisebox{.5pt}{\textcircled{\raisebox{-.9pt} {3}}}, and efficiency \raisebox{.5pt}{\textcircled{\raisebox{-.9pt} {4}}}. c, Distributions of the time (in seconds) to required to produce a final report for residents (top) and attendings (bottom). Asterisk (*) denotes statistical significance with a two-sided Mann-Whitney U test, $P < 0.0001$; \textit{n.s.} denotes differences that are not statistically significant. d, Evaluations on whether drafted reports answer the initial exam indication. Radiologists score reports on a five-point Likert scale ranging from -10 to 10. e, Opinions of radiologists on whether drafted reports improved their report writing and/or CXR interpretation efficiency.

%% file: 5_method.tex
\section*{\hspace{-0.5em}Method}
\subsection*{Description of the CheXinstruct dataset}
In this section, we describe our procedure for curating CheXinstruct, a large-scale training dataset consisting of 8.5 million samples.

\paragraph{Task Collection.}
An effective CXR FM must perform diverse interpretation and reasoning tasks. To this end, we first defined 35 tasks that users are likely to perform with CXR FMs (Fig.~\ref{fig:chexinstruct}a). Broadly, each task requires either (i) perception capabilities (\textit{i.e.,} the ability to understand the visual content of a CXR) or (ii) reasoning capabilities (\textit{i.e.,} the ability to make reasonable inferences or clinical decisions from a CXR). The 35 defined tasks come from five categories: (1) coarse-grained image perception tasks, which require the ability to understand CXRs as a whole (\textit{e.g.,} disease classification, view classification, and view matching); (2) fine-grained image perception, which require the ability to understand localized features in CXRs (\textit{e.g.,} abnormality detection, abnormality grounding, and foreign object detection); (3) text generation tasks, which require the ability to generate sections of radiology reports (\textit{e.g.,} findings generation, impression generation, and summarizing impressions from findings); (4) question answering tasks, which require the ability to respond to CXR-related questions (\textit{e.g.,} close-ended visual question answering (VQA), open-ended VQA, and difference VQA); and (5) miscellaneous tasks, which encompass other essential abilities for CXR FMs (\textit{e.g.,} image-text matching).

\paragraph{Source Dataset Collection.} To create training data samples for each of our 35 defined tasks, we first collected 32 publicly available datasets from diverse institutions: ChestXray14\supercite{wang2017chestxray14}, CheXpert\supercite{irvin2019chexpert,chambon2024chexpertplus}, MIMIC-CXR\supercite{johnson2019mimiccxr}, PadChest\supercite{bustos2020padchest}, RSNA\supercite{shih2019rsna}, COVIDX-CXR-3\supercite{pavlova2022covidx}, CXR-LT\supercite{holste2024cxrlt}, BRAX\supercite{reis2022brax}, NLM-TB\supercite{jaeger2014nlmtb}, MS-CXR-T\supercite{bannur2023mscxrt}, VinDr-CXR\supercite{nguyen2022vindr}, VinDr-PCXR\supercite{pham2022vindr}, Candid-PTX\supercite{feng2021candid}, SIIM\supercite{siim}, Object-CXR\supercite{objectcxr}, MS-CXR\supercite{boecking2022mscxr}, OpenI\supercite{demner2016openi}, BIMCV-COVID19\supercite{vaya2020bimcv}, ROCO\supercite{pelka2018roco}, MIMIC-III\supercite{johnson2016mimiciii}, VQA-RAD\supercite{lau2018vqarad}, SLAKE\supercite{liu2021slake}, MedVQA-2019\supercite{MedVQA2019}, PMC-VQA\supercite{zhang2023pmcvqa}, Rad-Restruct\supercite{pellegrini2023radrestruct}, MIMIC-CXR-VQA\supercite{bae2024mimiccxrvqa}, MIMIC-Diff-VQA\supercite{hu2023mimicdiffvqa}, RadQA\supercite{soni2022radqa}, ReXVal\supercite{yu2023rexval}, MIMIC-NLE\supercite{kayser2022mimicnle}, RadNLI\supercite{miura2021radnli}, and RadGraph\supercite{jain2021radgraph}. 
In total, we gathered 1,077,494 unique images (Fig.~\ref{fig:chexinstruct}b). Each image is paired with annotations, such as text (\textit{e.g.,} radiology reports or image captions), classification labels (\textit{e.g.} disease annotations), or visual grounding labels (\textit{e.g.} bounding boxes).

\paragraph{CheXinstruct Compilation.} 
To compile the CheXinstruct dataset, we first preprocessed the source datasets to ensure data quality. This process includes (1) manual quality control, where we randomly inspect examples from each dataset and design strategies to filter out low-quality or irrelevant samples (\textit{e.g.,} non-CXR images or noisy radiology reports), and (2) automated report restructuring, where we use a proprietary model (\textit{i.e.,} GPT-4) to impose structure on free-form radiology reports (Fig.~\ref{fig:chexinstruct}c). We also unified the diverse file and label structures of the source datasets. 
Next, we generated training data samples for each of our 35 defined tasks, with each sample consisting of a data triplet with an image, an instruction, and the desired response to the instruction (Fig.~\ref{fig:chexinstruct}d). For each task, we first selected source datasets with relevant annotations; for instance, when considering the task of disease classification, we selected datasets with disease classification labels. Then, for each image in the selected source dataset, we created an instruction by sampling from a list of ten manually-defined templates relevant for the task of interest. Instructions may be either multiple-choice questions, where we randomly sampled possible answer options, or open-ended queries. A response for the instruction was derived from the annotations associated with the image. In total, this process resulted in 8,466,352 data triplets with an instruction, at least one image, and a response (Fig.~\ref{fig:chexinstruct}e). We note that some tasks included in CheXinstruct are text-only (\textit{e.g.,} findings summarization), in which case no images are included in the triplet. We strictly followed the official or traditional dataset splits (training, validation, and test) to prevent data leakage.

\subsection*{Training CheXagent}
We then utilized CheXinstruct to train CheXagent, a CXR FM capable of processing images and instructions as input and generating free-form text responses as output. To this end, CheXagent consists of three core components: (1) an image encoder, which encodes images into low-dimensional features, (2) a vision-language projector, which projects visual features into the language representation space, and (3) a language decoder, which processes input instructions and visual features and generates output responses.

We began by training a language decoder (Fig.~\ref{fig:chexagent}a). Our goal in this stage was to create a language model with comprehensive medical and clinical knowledge. We adopted Phi-2\supercite{li2023phi2}, a 2.7 billion parameter decoder-only transformer model with 32 Transformer layers, each featuring 32 attention heads. We then trained the language decoder with data from four distinct sources: (1) clinical notes (\textit{e.g.,} discharge summary and radiology reports from MIMIC-IV), (2) scientific articles (\textit{e.g.,} PubMed Central articles), (3) Wikipedia-style text, and (4) general-domain text. To prevent data leakage, we excluded any studies from MIMIC-IV\supercite{johnson2023mimiciv} that were part of the validation and test sets of MIMIC-CXR. The total text corpus comprises 2,749,125,761 tokens. We used the causal language modeling (next-word prediction) loss to train the language decoder.

We then trained the image encoder to learn effective visual representations of CXRs (Fig.~\ref{fig:chexagent}b and c). We adopted SigLIP-Large\supercite{zhai2023siglip}, a transformer\supercite{vaswani2017transformer} model with 24 Transformer layers, each with 16 attention layers. SigLIP-Large was originally pretrained using the WebLi\supercite{chenpali} dataset. Here, we adapted this image encoder to the CXR domain. We first extracted image-report and image-caption pairs from the CheXinstruct dataset, resulting in 1,052,257 image-text pairs. We strictly adhered to the data split defined in CheXinstruct to avoid data leakage. We extended the input resolution of the model from 384 to 512 by interpolating the positional encodings. We then used the SigLIP loss function to train the image encoder using the collected image-text dataset.

After individually training the language decoder and image encoder, we developed a vision-language projector (a two-layer multi-layer perceptron) to project the visual features to the feature dimension of the language decoder (\textit{i.e.,} from 1,024 to 2,560) (Fig.~\ref{fig:chexagent}d). We trained this projector using the same set of 1,052,257 image-text pairs as the image encoder, with the image encoder and language model weights frozen. CheXagent was trained to generate reports or captions for each input image. Subsequently, we utilized the CheXinstruct dataset with (instruction, image, response) triplets to train CheXagent. CheXagent was trained to generate output responses given the images and instructions as input. We kept the image encoder unfrozen for one epoch and frozen for three epochs. We used the causal language modeling (next-word prediction) loss to train the language decoder. We detailed the training hyperparameters in Extended Data Table~\ref{table:hyperparameters}.


\subsection*{Building CheXbench}
We developed CheXbench, an evaluation benchmark for enabling systematic comparisons of FMs across 8 clinically-relevant CXR interpretation tasks (Fig.~\ref{fig:chexagent}e). CheXbench was structured with three evaluation axes, crafted to assess crucial aspects of CXR interpretation: (1) image perception, (2) image-text reasoning, and (3) text generation.

\paragraph{Image Perception.} We first evaluated the ability of FMs to understand the visual content of CXRs. We utilized the following three tasks, each formatted as an instruction with multiple choices:

\begin{enumerate}
    \item View Classification (600 samples): Given a CXR, the FM is tasked with identifying the imaging view. This is performed on the CheXpert (300 samples) and MIMIC-CXR (300 samples) test sets. Each instruction was associated with three multiple-choice options: anterior-posterior (AP), posterior-anterior (PA), or lateral.
    \item Temporal Classification (62 samples): Given two CXRs collected at different timepoints from a single patient, the FM is tasked with identifying the progression of a disease. This was performed using the MS-CXR-T dataset. Each instruction was associated with three multiple-choice options: improved, stable, or worsened. We considered five diseases: consolidation, edema, pleural effusion, pneumonia, and pneumothorax.
    \item Disease Identification (2,684 samples): We evaluated the ability of FMs to identify key findings in CXRs with the following three subtasks, which differed in the format of instructions:
    \begin{itemize}
        \item Binary Disease Classification (433 samples): Given a CXR, the FM is tasked with identifying whether a specific finding is present or absent in the image. We considered twelve findings from the CheXpert test set (annotated by expert radiologists), one finding (pneumonia) from the RSNA dataset, and one finding (pneumothorax) from the SIIM dataset. Each instruction was associated with two multiple-choice options: Yes and No. 
        \item Single Disease Identification (864 samples): Given a CXR, the FM is tasked with identifying a single finding present in the image. We considered 13 findings from the MIMIC-CXR test set, 13 findings from the CheXpert test set (annotated by expert radiologists), and 20 findings from OpenI (obtained from Medical Subject Heading (MeSH) codes). Instructions were associated with four options, each referencing a single finding (\textit{e.g.,} `pneumonia'). 
        \item Multi-Disease Identification (1,387 samples): Given a CXR, the FM is tasked with identifying a set of multiple findings present in the image. We again considered MIMIC-CXR, CheXpert, and OpenI. Instructions were associated with four options, each referencing a set of multiple findings (\textit{e.g.,} ``pneumonia, pleural effusion, cardiomegaly'').
    \end{itemize}
\end{enumerate}

We then provided the instruction and at least one image to the FM, and computed the accuracy of each FM in identifying the correct multiple-choice option within the generated response. We constructed each task to exhibit class balance to the extent possible.  

\paragraph{Image-Text Reasoning.} Next, we evaluated the ability of FMs to perform complex reasoning tasks on CXRs. We utilized the following three tasks:

\begin{enumerate}
    \item Fine-Grained Reasoning (380 samples): Given a CXR, the FM is tasked with differentiating between two subtly different findings. In contrast to single-disease classification, this task employed hard negatives, with each instruction associated with two challenging options distinguished by only a single word indicating the location or severity of a finding (\textit{e.g.,} ``left-sided pleural effusion'' vs. ``right-sided pleural effusion''). We implemented this task using the OpenI dataset.
    \item Visual-Question Answering (238 samples): We evaluated FMs across two standard VQA benchmarks: SLAKE and Rad-Restruct. Both SLAKE and Rad-Restruct consist of multiple-choice questions with two options: Yes and No.
    \item Phrase Grounding (149 samples): Given a CXR and a phrase, the FM is tasked with localizing the phrase to the corresponding region in the image. We implemented this task using the MS-CXR dataset. 
\end{enumerate}

For the fine-grained reasoning and VQA tasks, we utilized instructions with multiple choices; we then provided the instruction and a CXR to the FM, and computed the accuracy of each FM in identifying the correct multiple-choice option within the generated response. We constructed each task to exhibit class balance to the extent possible. For the phrase grounding task, we provided an open-ended instruction to the FM and evaluated the accuracy of the bounding box coordinates within the generated response.

\paragraph{Text Generation.} We evaluated the ability of FMs to generate and understand clinical text. We utilized the following two tasks, each formatted as an open-ended instruction:

\begin{enumerate}
    \item Findings Generation (2,451 samples): Given a CXR, the FM is tasked with generating the findings section of the radiology report, identifying critical features such as the presence of abnormalities. We implemented this task using the MIMIC-CXR and CheXpert datasets.
    \item Findings Summarization (1,394 samples): Given the findings section of a radiology report, the FM is tasked with summarizing the key observations into a concise statement, referred to as the impressions section. We note that this task is text-only and does not include images. We implemented this task using MIMIC-CXR.
\end{enumerate}

We provided the instruction and a CXR to the FM, and evaluated the quality of the generated free-form response with standard natural language evaluation metrics.

On the tasks of View Classification, Temporal Classification, Disease Identification, Fine-Grained Reasoning, and Visual-Question Answering, we compared CheXagent with one general-domain instruction-tuned FM (QwenVL\supercite{bai2023qwen}), two medical-domain FMs (LLaVA-Med\supercite{li2024llavamed} and RadFM\supercite{wu2023radfm}), and one proprietary model (GPT-4\supercite{achiam2023gpt4}). In Extended Data Fig.~\ref{fig:evaluation-extended-data}, we also compared CheXagent with BLIP-2\supercite{li2023blip2}, InstructBLIP\supercite{dai2023instructblip}, MedFlamingo\supercite{moor2023med}, and XrayGPT\supercite{thawkar2023xraygpt}. We reported accuracy as our evaluation metric. On the task of Phrase Grounding, we compared CheXagent with two zero-shot contrastive models (BioVIL\supercite{bannur2023biovil} and CheXzero\supercite{tiu2022chexzero}), one single-task supervised visual grounding model (TransVG\supercite{deng2021transvg}), and one multi-task supervised model (ChEX\supercite{muller2024chex}). On the task of Findings Generation, we compared four medical-domain FMs (MedFlamingo\supercite{moor2023med}, LLaVA-Med\supercite{li2024llavamed}, RadFM\supercite{wu2023radfm}, and XrayGPT\supercite{thawkar2023xraygpt}) and three proprietary models (GPT-4V\supercite{achiam2023gpt4}, Med-PaLM-M\supercite{tu2024medpalmm}, and MAIRA-1\supercite{hyland2023maira}) using three text domain-specific and semantic similarity evaluation metrics (CheXbert-F1, BERTScore, and RadGraph-F1). On the task of Findings Summarization task, we compared CheXagent with four large language models (\textit{i.e.,} LLaMA\supercite{touvron2023llama}, Vicuna\supercite{vicuna2023}, FLAN-T5-XL\supercite{chung2024flan}, and FLAN-UL2\supercite{chung2024flan}) specifically adapted to MIMIC-CXR, similar to an existing study\supercite{van2024adapted}. We reported ROUGE-L\supercite{lin2004rouge}, a classic summarization metric, for this task.

\subsection*{Setup of the Reader Study for Clinical Evaluation}
To complement automated quantitative evaluation, we also conducted a qualitative expert reader study to evaluate the potential clinical efficacy benefits of CheXagent in real-world practice. Our study mimicked the typical workflow seen in real-world academic radiology departments, where radiology residents draft initial reports and attending radiologists review for accuracy and make necessary edits. Our study evaluated the role of CheXagent in drafting the initial reports. In particular, we evaluated the utility of CheXagent across two axes: (1) efficiency (\textit{i.e.,} whether using CheXagent-drafted reports can improve radiologist efficiency), and (2) accuracy (\textit{i.e.,} whether CheXagent-drafted reports are high-quality in nature).

To this end, our readers included four radiology residents and four attending radiologists. For resident radiologists, we considered two settings: (1) writing reports from scratch for 10 cases and (2) editing CheXagent-drafted reports for 20 cases. For attending radiologists, we also considered two settings: (1) editing resident-drafted reports for 10 cases and (2) editing CheXagent-drafted reports for 20 cases. To ensure a diverse selection of CXR studies, we randomly sampled 50 cases from the test set of MIMIC-CXR and distributed 30 cases to each reader. We deployed this reader study via a user interface created using Streamlit\footnote[4]{https://streamlit.io/}.

We collected the following metrics and feedback from our reader study:
\begin{enumerate}
    \item \textit{Time required to produce a report}. We used our Streamlit application to automatically record the time (in seconds) taken to write a radiology report for each case. For cases where a CheXagent-drafted report or a resident-written report was provided to a reader, we pre-filled the submission textbox with the drafted report and prompted the reader to make edits. For cases where the reader was required to write a report from scratch, a blank textbox was provided.
    \item \textit{Applicability of report to exam indication}: We asked readers to rate whether a provided draft report addresses the exam indication on a five-point Likert scale (weighted from -10 to 10 during analysis).
    \item \textit{Reasons for editing}: We prompted readers to explain their reasoning for making edits to drafted reports. We offered a list of options (grouped into `content' and `style') that the readers could select to explain their reasoning for edits. The durations for providing these feedback responses were not included in the report generation efficiency computation described above.
    \item \textit{Efficiency feedback}: We asked readers if the drafted report (either from CheXagent or residents) improved their efficiency in writing and/or interpretation. Readers responded with a Yes or No answer.

\end{enumerate}
Textboxes were provided to collect qualitative feedback. To avoid distracting the readers, the feedback section was shown only after the readers finished editing reports and clicked a submit button.

\subsection*{Statistics and reproducibility}
We computed 95\% confidence intervals using bootstrapping with 1,000 samples with replacement for all CheXagent analyses. A two-sided paired t-test was used to evaluate the statistical significance of performances between the best and second-best models for each task. All results by CheXagent were obtained using greedy sampling with the beam size set to 1, ensuring reproducibility. For the clinical reader study, a two-sided Mann-Whitney U test was used to evaluate the statistical significance of differences between different reader study settings. Samples in the reader study were displayed in a random order, and the readers were blinded to the source of the drafted reports (either from CheXagent or radiologists).

\subsection*{Data Availability}
This study utilized datasets that are publicly accessible. Those requiring Physionet access due to their terms of use have references provided in the manuscript. For other datasets not requiring Physionet access, researchers can access the original versions through manuscript references. The CheXinstruct dataset, the model weights of different stages, and the CheXbench evaluation benchmark will be released before publication.

\subsection*{Code Availability}
The code used for experiments in this study will be made publicly available before publication. We built upon the open-source libraries PyTorch and Transformers. It includes the preprocessing script to curate CheXinstruct, the code to train CheXagent, the CheXbench evaluation scripts for existing FMs and CheXagent, and the interface implementation of the clinical reader study. All the models will be hosted on HuggingFace (https://huggingface.co/) before publication.

\subsection*{Acknowledgements}
A.S.C. receives research support from the National Institutes of Health (grants - R01 HL167974, R01 AR077604, R01 EB002524, R01 AR079431, P41 EB027060, and contracts 75N92020C00008, 75N92020C00021); and from GE Healthcare, Philips, Amazon, Microsoft/OpenAI, and Stability.ai. C.B. receives research support from the Promedica Foundation, Chur, Switzerland. Research reported in this publication was made possible in part by the National Institute of Biomedical Imaging and Bioengineering (NIBIB) of the National Institutes of Health which supports the Medical Imaging and Data Resource Center under contracts 75N92020C00008 and 75N92020C00021, and by grant \#1R18HS028955 from the Agency for Health Research and Quality.

\subsection*{Author contributions}
Z.C. and M.V. designed the study and carried out the data collection, data analysis, model construction, and benchmark design. Z.C., M.V., M.P., D.V.V., and J.B.D. carried out the technical model evaluation. A.S.C., S.G., D.V.V., Z.C., J.X., M.V., J.B.D., and C.P.L. designed the clinical reader study. J.X. and Z.C. implemented the reader study. J.X., Z.C., M.V., A.Y., C.O., A.J., S.A., M.S.E.M., E.P.R., E.B.T., C.B., C.F.B, and S.G. carried out the reader study and interpreted the results. Z.C., M.V., J.X., M.P., D.V.V., A.Y., C.B., L.B., J.M.J.V., E.P.R., J.P.C., T.M.A, J.J., J.B.D., A.S.C., and C.P.L. contributed to the technical discussions. All authors contributed to the drafting and revision of the manuscript. J.B.D., A.S.C., and C.P.L. supervised and guided the research.

%% file: 6_appendix.tex
\clearpage
\captionsetup[figure]{labelformat=default, labelsep=colon, name=Extended Data Figure}
\captionsetup[table]{labelformat=default, labelsep=colon, name=Extended Data Table}
\renewcommand{\thefigure}{\arabic{figure}}
\setcounter{figure}{0}
\setcounter{table}{0}

\section*{\hspace{-0.5em}Extended Data}
\begin{figure}[h]
\centering
\includegraphics[width=0.99\textwidth, trim=0 0 0 0]{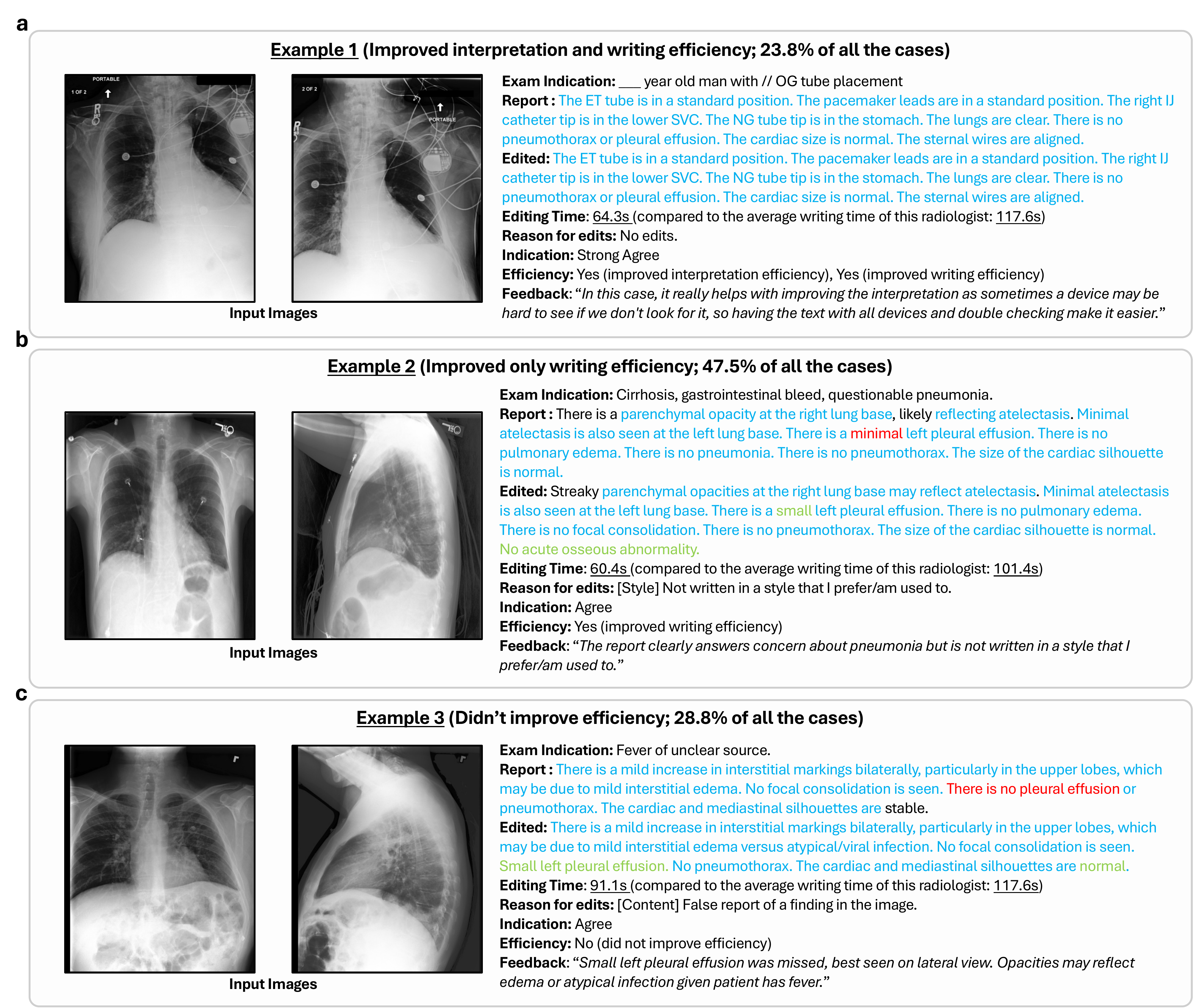}
\caption{\textbf{Qualitative analysis of three cases from the reader study}. Blue text represents accurate findings in CheXagent-drafted reports, red text represents false predictions in CheXagent-drafted reports, and green text represents findings missed by CheXagent. a, An example case where a radiologist found the CheXagent-drafted report to improve both interpretation and writing efficiencies. Here, CheXagent identified all four devices in the CXR study, enabling the radiologist to efficiently generate the final report. b, An example case where a radiologist found the CheXagent-drafted report to improve writing efficiency. Here, CheXagent accurately predicts the majority of the findings, and the radiologist reorganized and edited the report in his preferred style. c, An example case where a radiologist found the CheXagent-drafted report to not improve efficiency. Here, CheXagent missed a finding (left pleural effusion) in the CXR study.}
\label{fig:reader-study-case-study}
\end{figure}

\begin{figure}[!t]
\centering
\includegraphics[width=0.99\textwidth, trim=0 0 0 0]{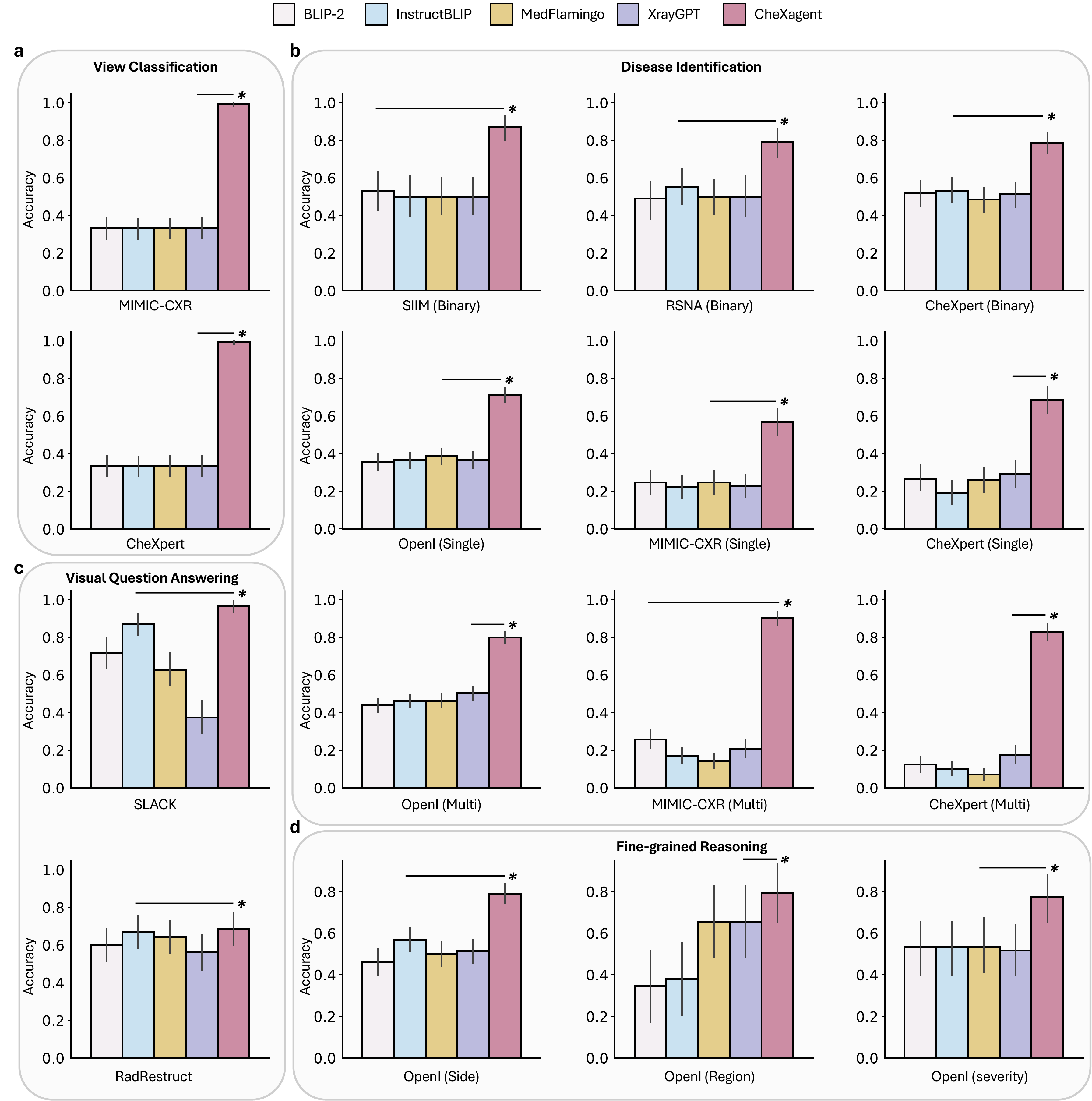}
\caption{\textbf{Technical evaluation on more FMs.} We compared CheXagent with BLIP-2\supercite{li2023blip2}, InstructBLIP\supercite{dai2023instructblip}, MedFlamingo\supercite{moor2023med}, and XrayGPT\supercite{thawkar2023xraygpt}. a, Performance of FMs on view classification. Bar graphs show mean accuracy with 95\% confidence intervals. b, Performance of FMs on disease identification with three subtasks.  Bar graphs show mean accuracy with 95\% confidence intervals. c, Performance of FMs on visual question answering. The bar graph shows mean accuracy with 95\% confidence intervals. d, Performance of FMs on fine-grained reasoning. Bar graphs show mean accuracy with 95\% confidence intervals.}
\label{fig:evaluation-extended-data}
\end{figure}

\input{hyperparameters}

%% file: hyperparameters.tex
\begin{table}[t]
\resizebox{0.95\linewidth}{!}{
\begin{tabular}{@{}lcccc@{}}
\toprule
\multirow{2}{*}{\textbf{Configuration}} & \textbf{Language Model}           & \textbf{Vision-Language} & \textbf{Instruction Tuning}          & \multirow{2}{*}{\textbf{Instruction Tuning}} \\
                                        & \textbf{(Continued) Pre-training} & \textbf{Pre-training}    & \textbf{(Vision-Language Alignment)} &                                              \\ \midrule
ViT Init.                               & -                                 & SigLIP-Large             & from Stage 2                         & from Stage 3                                 \\
LLM Init.                               & Phi-2                             & -                        & from Stage 1                         & from Stage 3                                 \\
VL Projector init.                      & -                                 & -                        & random                               & from Stage 3                                 \\
Image Resolution                        & -                                 & $518^{2}$                & $518^{2}$                            & $518^{2}$                                    \\
ViT sequence length                     & -                                 & 1,024                    & 1,024                                & 1,024                                        \\
LLM sequence length                     & 4,096                             & -                        & 4,096                                & 4,096                                        \\
Optimizer                               & \multicolumn{4}{c}{AdamW}                                                                                                                          \\
Optimizer hyperparameter                & \multicolumn{4}{c}{$\beta_{1} = 0.9$, $\beta_{2} = 0.98$, $eps = 1e-6$}                                                                            \\
Peak learning rate                      & 2e-5                              & 5e-4                     & 1e-4                                 & 1e-5                                         \\
Learning rate schedule                  & \multicolumn{4}{c}{cosine decay}                                                                                                                   \\
Weight decay                            & 0.1                               & 0.2                      & 0.1                                  & 0.1                                          \\
Gradient clip                           & \multicolumn{4}{c}{1.0}                                                                                                                            \\
Training epochs                         & 3                                 & 20                       & 3                                    & 4                                            \\
Warm-up ratios                          & 0.05                              & 0.05                     & 0.05                                 & 0.05                                         \\
Global batch size                       & 1,024                             & 512                      & 512                                  & 256                                          \\
Gradient Acc.                           & \multicolumn{4}{c}{1}                                                                                                                              \\
Numerical precision                     & \multicolumn{4}{c}{bfloat16}                                                                                                                       \\
DeepSpeed                               & ZoRO-2                            & -                        & ZoRO-2                               & ZoRO-3                                       \\ \bottomrule
\end{tabular}}
\caption{Training hyperparameters of CheXagent.}
\label{table:hyperparameters}
\end{table}